\documentclass[acmtog]{acmart}
\acmSubmissionID{274}

\usepackage{booktabs} 
\usepackage{bm}

\usepackage[ruled]{algorithm2e} 

\SetAlFnt{\small}
\SetAlCapFnt{\small}
\SetAlCapNameFnt{\small}
\SetAlCapHSkip{0pt}

\acmJournal{TOG}
\acmYear{2023} 
\acmVolume{42} 
\acmNumber{6} 
\acmArticle{} 
\acmMonth{12} 
\acmPrice{15.00}
\acmDOI{10.1145/3618333}

\setcopyright{rightsretained}

\definecolor{MyDarkGreen}{RGB}{45,155,45}

\begin{document}
\title{Object Motion Guided Human Motion Synthesis}


\author{Jiaman Li}
\affiliation{%
 \institution{Stanford University}
 \country{USA}}
\email{jiamanli@stanford.edu}
\author{Jiajun Wu\textsuperscript{$\dagger$}}
\affiliation{%
 \institution{Stanford University}
 \country{USA}
}
\email{jiajunwu@cs.stanford.edu}
\author{C. Karen Liu\textsuperscript{$\dagger$}}
\affiliation{%
\institution{Stanford University}
\country{USA}}
\email{karenliu@cs.stanford.edu}

\begin{abstract}
\renewcommand{\thefootnote}{\fnsymbol{footnote}}
\footnotetext[2]{indicates equal contribution.}
Modeling human behaviors in contextual environments has a wide range of applications in character animation, embodied AI, VR/AR, and robotics. In real-world scenarios, humans frequently interact with the environment and manipulate various objects to complete daily tasks. In this work, we study the problem of full-body human motion synthesis for the manipulation of large-sized objects. We propose \textbf{O}bject \textbf{MO}tion guided human \textbf{MO}tion synthesis (OMOMO), a conditional diffusion framework that can generate full-body manipulation behaviors from only the object motion. Since naively applying diffusion models fails to precisely enforce contact constraints between the hands and the object, OMOMO learns two separate denoising processes to first predict hand positions from object motion and subsequently synthesize full-body poses based on the predicted hand positions. By employing the hand positions as an intermediate representation between the two denoising processes, we can explicitly enforce contact constraints, resulting in more physically plausible manipulation motions. With the learned model, we develop a novel system that captures full-body human manipulation motions by simply attaching a smartphone to the object being manipulated. Through extensive experiments, we demonstrate the effectiveness of our proposed pipeline and its ability to generalize to unseen objects. Additionally, as high-quality human-object interaction datasets are scarce, we collect a large-scale dataset consisting of 3D object geometry, object motion, and human motion. Our dataset contains human-object interaction motion for 15 objects, with a total duration of approximately 10 hours. 

\end{abstract}



\begin{teaserfigure}
  \includegraphics[width=\textwidth]{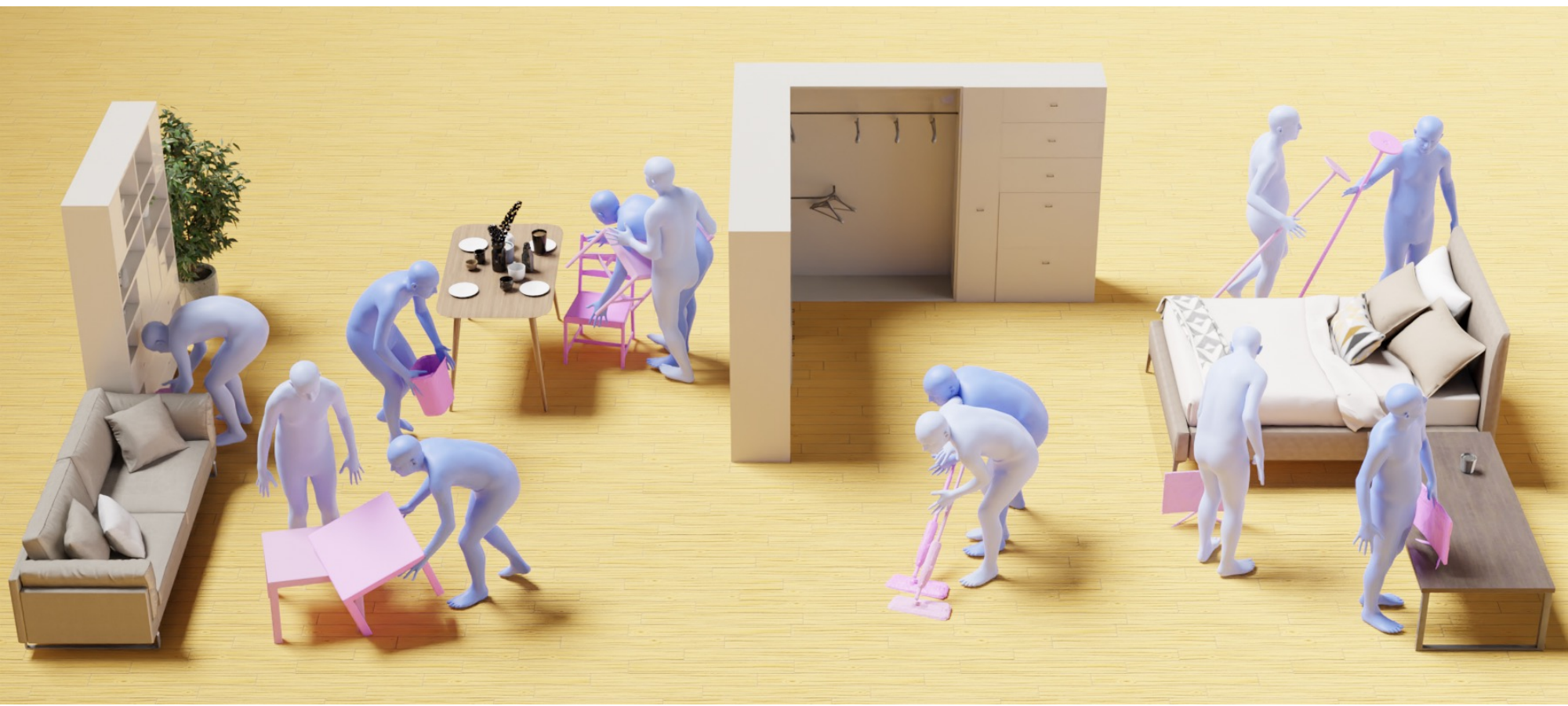}
  \vspace{-7mm}
  \caption{OMOMO takes a sequence of object states as input and generates full-body human motion interacting with the given object. 
  }
  \label{fig:teaser}
\end{teaserfigure}

\maketitle

\section{Introduction}
Capturing and synthesizing human movements in contextual environments is critical to progressing embodied AI, character animation, VR/AR, and robotics. The real world in which humans live is complex and highly dynamic. Humans routinely interact with dynamic objects to accomplish everyday tasks, demonstrating a diverse range of full-body manipulations. For example, humans pull and push a mop to tidy a floor, reposition a floor lamp to illuminate a specific area, drag a chair toward a desk, and place a monitor on a desk. Realistically simulating such complex manipulation behaviors is a fundamental problem in computer graphics with a lot of downstream applications. 

Prior works have made significant progress in addressing the contextual human motion synthesis problem for activities such as navigating through a 3D scene or sitting on a chair~\cite{wang2021synthesizing,wang2021scene,hassan_samp_2021,zhang2022couch,Zhao:ICCV:2023,mir2023generating}. They model interactions with static 3D scenes or static objects based on large-scale human motion datasets. In comparison, datasets containing full-body interaction with moving objects are scarce. Prior works rely on reinforcement learning to model such behaviors \cite{merel2020catch,hassan2023synthesizing,xie2023hierarchical}, but the learned policies are often limited to manipulating specific types of geometry used for training.

We present a new approach to synthesizing the dynamic interactions between humans and large-sized objects, particularly in manipulation tasks requiring full-body movements and precise coordination between hands and objects. We aim to bridge the gap between current research and real-world manipulation behaviors by introducing a large-scale dataset and developing a robust approach to synthesize full-body motion from object motion.  

We present a new framework -- \textbf{O}bject \textbf{MO}tion guided human \textbf{MO}tion synthesis (OMOMO). We leverage a conditional diffusion formulation to predict plausible full-body poses with a sequence of object geometry as input. One key observation is that hand position is a deciding factor for full-body movement during manipulation. Thus, we devise a two-stage approach to generate hand positions conditioned on object geometry features and then synthesize full-body poses based on the predicted hand positions. The two-stage design enables us to apply contact constraints to our predicted hand joint positions, which significantly enhances the contact realism of the generated results. We demonstrate the effectiveness of our proposed method in our dataset and showcase its generalizability to unseen objects. 

Moreover, we introduce an innovative application that generates full-body human poses based on object motion captured by an iPhone. In particular, we mount an iPhone on an unseen object, employ the iPhone ARKit to obtain camera poses and deduce the motion of the object. Subsequently, we apply these object poses to 3D geometry reconstructed using Luma~\cite{lumaai}. Our pipeline takes the sequence of object geometry as input and generates the corresponding full-body human motion. This application demonstrates an affordable and user-friendly method for capturing human interaction motions during everyday tasks. 

An additional contribution of this work is a new dataset with paired object motion and human motion to facilitate the learning of full-body human manipulation behaviors. We leverage an advanced 3D reconstruction technique to extract 3D object geometry from a monocular video. We then use motion capture devices to capture human and moving objects simultaneously. To capture motions that resemble real-world scenarios, we provide language descriptions to guide our volunteers to perform meaningful interactions with various objects. Our dataset can be used for different tasks to model full-body human manipulation behaviors. 

To summarize, the contributions of this work include:
\begin{enumerate}
\vspace{-10pt}
    \item A novel approach to full-body manipulation synthesis by generating full-body motion from object motion. We introduce an effective framework based on conditional diffusion to synthesize full-body movements from object motion.
    \item A novel application that employs an iPhone to capture object motion from the egocentric view of the object, enabling the synthesis of full-body movements by simply attaching an iPhone to various objects. 
    \item A large-scale high-quality dataset consisting of 3D object geometry, object motion, and full-body motion.
\end{enumerate}
\vspace{10pt}

\section{Related Work}

\paragraph{Human Motion and Interaction Datasets.}

Human motion modeling has been extensively studied with motion capture datasets~\cite{AMASS}. Recently, there has been a surge of interest in human scene interactions. PROX~\cite{prox} provides paired 3D scenes and human motions extracted from RGB videos. HPS~\cite{HPS} contributes a dataset of paired scenes, egocentric video, and human motion captured with an IMU-based suit. EgoBody~\cite{zhang2022egobody} collects a dataset consisting of 3D scenes, egocentric video, eye gaze, and human motions extracted from multi-view RGBD frames with a focus on social interactions.  GIMO~\cite{zheng2022gimo} explores the problem of gaze-guided motion prediction using a similar data modality. Synthetic datasets~\cite{li2022ego, wang2022humanise} combine scene datasets~\cite{replica19arxiv,dai2017scannet} with motion datasets~\cite{AMASS} to produce paired human motions in 3D environments. CIRCLE~\cite{araujo2023circle} integrates VR and MoCap techniques to collect high-quality motion within virtual scenes. 

A couple of datasets focus on human-object interactions. For example, SAMP~\cite{hassan_samp_2021} contains sitting and lying down motions while interacting with chairs and sofas. COUCH~\cite{zhang2022couch} is dedicated to data collection for sitting on different chairs. These datasets primarily contain motions interacting with static objects. 

Moreover, some datasets collect both human motion and object motion~\cite{GRAB:2020,bhatnagar22behave,guzov23ireplica,fan2023arctic}. GRAB~\cite{GRAB:2020} focuses on the interaction between humans and small-sized objects, involving mostly hand motions. BEHAVE~\cite{bhatnagar22behave} records interactions with larger-sized objects, making it closely related to our dataset. However, it relies on multi-view RGBD input to extract human and object motion, which does not yield motion of sufficient quality for motion synthesis tasks. Furthermore, the limited data for each object impedes its capacity for training a motion generative model. In contrast, our work focuses on synthesizing dynamic human interactions with large-sized objects and we introduce a large-scale dataset consisting of high-quality human motion and object motion. 

\paragraph{Contextual Human Motion Synthesis.} 
Motion synthesis is a long-standing problem in computer graphics, and here we survey prior works centered on motion synthesis in 3D environments. Leveraging the dataset with paired scenes and human motions~\cite{prox}, a couple of work~\cite{wang2021scene,wang2021synthesizing} learn separate modules to predict root trajectory first and generate full-body poses conditioned on both scene and the planned path. However, constrained by the scale and motion quality of the dataset, these methods struggle to synthesize realistic human motions. To improve the motion quality of generation results, SAMP~\cite{hassan_samp_2021} collects a high-quality dataset consisting of walking, sitting and lying down motions. And they present a pipeline that first produces a collision-free path based on A$^*$ algorithm, generates full-body motion following the path and then synthesizes interaction motions to sit on chairs and sofas. A recent work~\cite{mir2023generating} introduces action keypoints as scene abstraction, enabling continual motion synthesis generation across various scenes. In order to produce physically plausible movements, several works employ reinforcement learning techniques to learn interaction policies through meticulously designed task rewards~\cite {hassan2023synthesizing,lee2023locomotion,chao2021learning}. 

Another line of work focuses on reaching motion synthesis within contextual environments. GOAL~\cite{taheri2022goal} and SAGA~\cite{wu2022saga} generate full-body poses aimed at grasping a specific object. IMoS~\cite{ghosh2022imos} further synthesizes human and object motions simultaneously after grasping an object. However, these works only consider the target object and do not involve navigation in cluttered scenes. Meanwhile, CIRCLE~\cite{araujo2023circle} incorporates human-scene interaction features and formulates the problem with a scene-aware motion refinement model, enabling reaching synthesis in complex static scenes.

While most existing work that involves interaction with dynamic objects aims to synthesize dexterous hand motions~\cite{ye2012synthesis,li2007data,zhang2021manipnet}, our work diverges from this line of work. Instead, we focus on the synthesis of full-body movements for manipulation without synthesizing detailed hand movements.

Full-body human motion synthesis for manipulation has been explored in both kinematic-based~\cite{starke2019neural} and physics-based methods~\cite{hassan2023synthesizing, merel2020catch,xie2023hierarchical}.
NSM~\cite{starke2019neural} learns a gating network and a motion prediction network to synthesize interaction movements including sitting and carrying objects. As for physics-based character animation, reinforcement learning has been widely used to learn different skills~\cite{peng2018deepmimic,peng2021amp,xie2022learning,liu2018learning}. In terms of manipulation, ~\citet{merel2020catch} devise a hierarchical reinforcement learning framework to synthesize box catching and carrying movements with egocentric observations. More recently, ~\citet{hassan2023synthesizing} propose to learn policies based on the Adversarial Motion Priors framework~\cite{peng2021amp} for box manipulation task. 

In summary, most prior research has not considered the dynamic interaction between humans and large-sized objects. A few works studied the problem of full-body manipulation but were constrained to interactions with boxes. In contrast, our work examines contextual environments with diverse dynamic objects. Leveraging our large-scale dataset, we develop an approach to synthesize manipulation movements for diverse objects. And inspired by the success of diffusion in motion modeling~\cite{li2022ego,dabral2022mofusion,tevet2022human,zhang2022motiondiffuse,tseng2022edge,huang2023diffusion}, we design our framework based on conditional diffusion.

\section{Method}

\begin{figure*}[ht!]
    \includegraphics[width=7.0in]{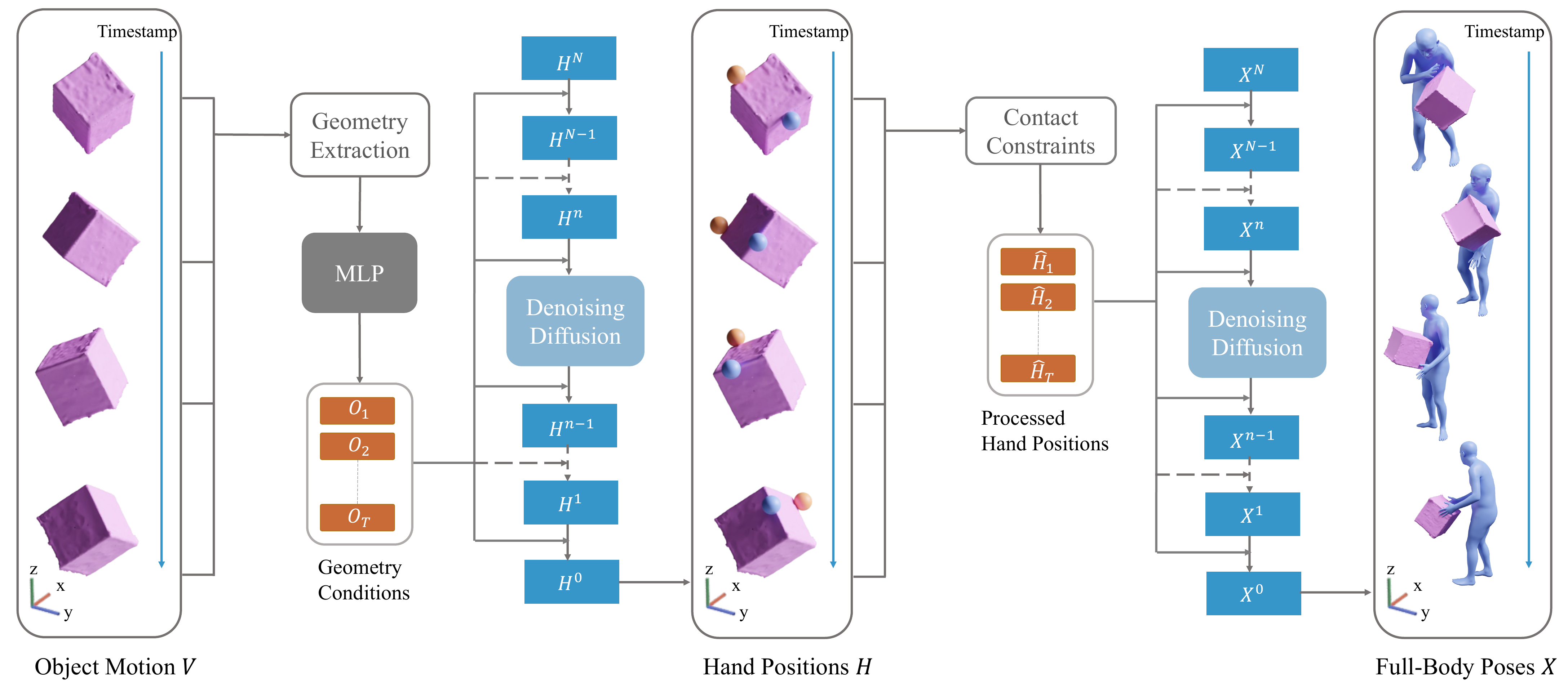}
    \vspace{-4mm}
    \caption{Method Overview. Given a sequence of object geometry, we use BPS representation to encode geometry features and project the representation to a low dimensional vector at each time step using an MLP. We use conditional diffusion to synthesize hand joint positions and apply contact constraints. Then we feed the updated hand joint positions to our full-body synthesis module and produce human poses in contact with the given dynamic object.}
    \label{fig:method}
\end{figure*}

Our goal is to generate full-body poses $\bm{X} \in \mathbb{R}^{T \times D}$ from a sequence of object geometry $\bm{V} \in \mathbb{R}^{T \times K \times 3}$, where $T$ denotes the time steps of the sequence, $D$, $K$ represents the dimension of human pose state and the number of vertices on object mesh respectively. This problem presents two significant challenges. First, there is inherent uncertainty in predicting full-body poses from object motions, as humans can produce the same object motion with varying movements. Second, the generated human poses need to maintain correct contact with the given object when it is being manipulated. 
The first challenge can be addressed by using a generative model, such as a diffusion model~\cite{ho2020denoising}. However, naively applying
diffusion models would not address the second challenge of precisely enforcing contact constraints between the hands and the object. We develop a two-staged method based on a diffusion framework with hand positions as an intermediate representation. The first stage predicts both right and left hand positions $\bm{H} \in \mathbb{R}^{T \times 6}$ from the object geometry. The second stage generates full-body poses $\bm{X} \in \mathbb{R}^{T \times D}$ conditioned on the predicted hand joint positions. Our pipeline is shown in Figure~\ref{fig:method}.

\subsection{Data Representation}
\paragraph{Human Pose Representation}
Our pose state representation at time step $t$ consists of global joint position $\bm{J}_t \in \mathbb{R}^{24 \times 3}$ and global joint rotation $\bm{Q}_t \in \mathbb{R}^{22 \times 6}$ represented using 6D continuous rotation~\cite{zhou2019continuity}. We adopt a widely used parametric human model, SMPL-X~\cite{smplx}, to reconstruct human mesh from pose and shape parameters. 

\paragraph{Object Representation}
Given a sequence of object geometry $\bm{V} \in \mathbb{R}^{T \times K \times 3}$, we adopt Basis Point Set (BPS) representation~\cite{prokudin2019efficient} to encode object geometry. We use the BPS representation for two reasons. First, it gives us a lightweight and compact representation using fixed-length vectors. Second, BPS does not rely on special model architecture, such as PointNet~\cite{qi2017pointnet}, to process and can be encoded with an MLP to learn downstream tasks effectively as demonstrated in the previous work~\cite{prokudin2019efficient}. We define a ball with a radius $r = 1$, a value chosen to encompass all objects in our dataset. The ball is centered at the centroid of the object, $(g^x_t, g^y_t, g^z_t) = \frac{1}{K}\sum_{i=1}^{K}\bm{V}_t^i$ at time step $t$. We sample 1024 points from the volume of the ball $\bm{B}_t \in \mathbb{R}^{1024 \times 3}$. The BPS representation is computed by calculating the difference between each sampled point and its nearest neighbor vertex on the object mesh, and denoted as $d(\bm{B}_t, \bm{V}_t) \in \mathbb{R}^{1024 \times 3}$. 
As the global position is not encoded in the BPS representation, we concatenate the 3D location of the object at time step $t$ to yield object geometry features $[g^x_t, g^y_t, g^z_t, d(\bm{B}_t, \bm{V}_t)] \in \mathbb{R}^{3+1024\times 3}$. 
Then we employ a Multilayer Perceptron (MLP) to project the high-dimensional features onto a lower-dimensional space. The projected geometry features are denoted as $\bm{O}_{t}, \bm{O}_t \in \mathbb{R}^{256}$. 

\subsection{Conditional Diffusion Formulation}

The diffusion model consists of a forward diffusion process and a reverse diffusion process. The forward diffusion process is gradually adding noise to the data representation $\bm{x}^0$ for $N$ steps formulated using a Markov chain, 
\begin{equation}
\label{eq:forward_diffusion}
    q(\bm{x}^{1:N}|\bm{x}^{0}) := \prod_{n=1}^{N} q(\bm{x}^{n}|\bm{x}^{n-1}).
\end{equation}
The transition of forward diffusion is modeled by a posterior distribution $q$. And each step is decided by a fixed variance schedule using $\beta_{n}$ and is defined as
\begin{equation}
\label{eq:forward_diffusion_step}
    q(\bm{x}^{n}|\bm{x}^{n-1}) := \mathcal{N}(\bm{x}^{n}; \sqrt{1-\beta_{n}}\bm{x}^{n-1}, \beta^{n}\bm{I}),
\end{equation}
where $\bm{I}$ represents identity matrix.

\begin{figure}[t]
\begin{center}
    \includegraphics[width=.9\columnwidth]{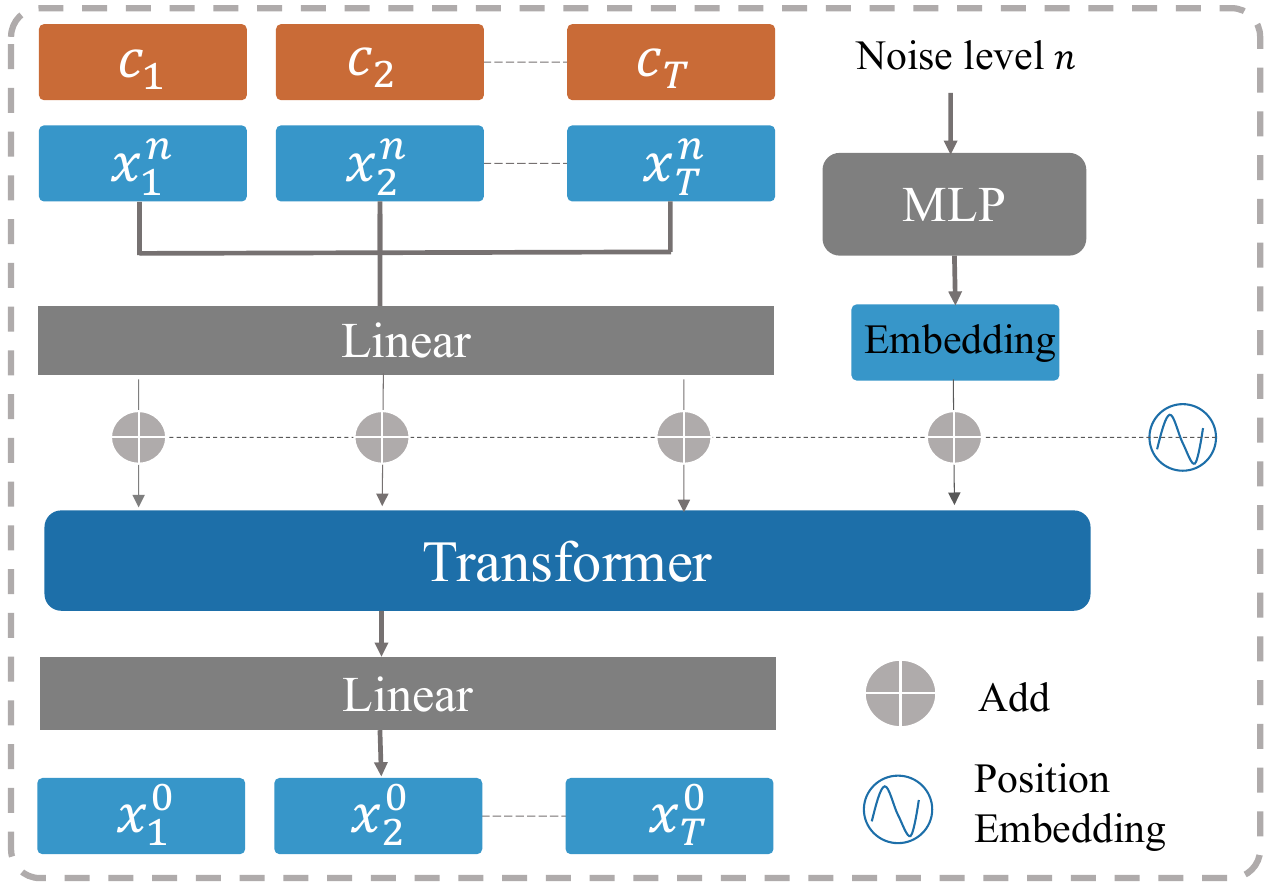}
    \vspace{-3mm}
    \caption{Model architecture of denoising network. In stage 1, the conditions $\bm{c}$ are object geometry features $\bm{O}$, and $\bm{x}$ are hand joint positions $\bm{H}$. In stage 2, the conditions $\bm{c}$ are rectified hand joint positions $\hat{\bm{H}}$, and $\bm{x}$ are full-body human poses $\bm{X}$.}
    \label{fig:architecture}
   \vspace{-5mm}
\end{center}
\end{figure}

The reverse diffusion process is to generate desired data representation from random noise $\bm{x}^N \sim \mathcal{N}(0,\bm{I})$. This is achieved by learning a neural network $p_{\theta}$ to denoise recursively. Specifically, at noise level $n$, we use $\bm{c}$ to represent the conditions, and we have the reverse diffusion process represented as follows:
\begin{equation}
\label{eq:reverse_diffusion_step}
    p_{\theta}(\bm{x}^{n-1}|\bm{x}^{n}, \bm{c}) := \mathcal{N}(\bm{x}^{n-1}; \bm{\mu}_{\theta}(\bm{x}^n, n, \bm{c}), \sigma_{n}^{2}\bm{I}),
\end{equation}
where $\bm{\mu}_{\theta}(\bm{x}^n, n, \bm{c})$ is the learned mean, $\sigma_{n}$ is the fixed variance. 
$\bm{\mu}_{\theta}(\bm{x}^n, n, \bm{c})$ (we use $\bm{\mu}_{\theta}$ in the following equation for brevity) can be formulated as, 
\begin{equation}
\label{eq:reverse_diffusion_mu}
     \bm{\mu}_{\theta} = \frac{\sqrt{\alpha_{n}}(1-\bar{\alpha}_{n-1})\bm{x}^{n}+\sqrt{\bar{\alpha}_{n-1}}(1-\alpha_{n})\hat{\bm{x}}_{\theta}(\bm{x}^{n}, n, \bm{c})}{1-\bar{\alpha}_{n}},
\end{equation}
where $\hat{\bm{x}}_{\theta}(\bm{x}^{n}, n, \bm{c})$ is the prediction of $x^0$, $\alpha_{n}, \bar{\alpha}_{n}$ are fixed parameters that satisfy $\bar{\alpha}_{n} = \prod_{i=1}^{n} \alpha_{n}$.

Learning the mean can be reparameterized as learning to predict the original data $\bm{x}^0$. We use reconstruction loss of $\bm{x}^{0}$ during training:
\begin{equation}
\label{eq:loss}
     \mathcal{L} = \mathbb{E}_{\bm{x}^0, n}||\hat{\bm{x}}_{\theta}(\bm{x}^{n}, n, \bm{c}) - \bm{x}^{0}||_{1}.
\end{equation}

\subsection{Our Pipeline}
\paragraph{Generating Hand Positions from Object Geometry.}
In the first stage, we employ conditional diffusion to generate hand joint positions $\bm{H}_1, \bm{H}_2, ..., \bm{H}_{T}$ from object geometry features $\bm{O}_1, \bm{O}_2, ..., \bm{O}_{T}$. Here, the conditions $\bm{c}$ are represented by $\bm{O}_1, \bm{O}_2, ..., \bm{O}_T$. We adopt a transformer model architecture~\cite{vaswani2017attention} as our denoising network which consists of four self-attention blocks. Each self-attention block contains a multi-head attention layer followed by a position-wise feedforward layer. As shown in Figure~\ref{fig:architecture}, we introduce an additional step to include noise level embedding as an input to our transformer model.

\begin{figure*}[ht!]
    \includegraphics[width=7.1in]{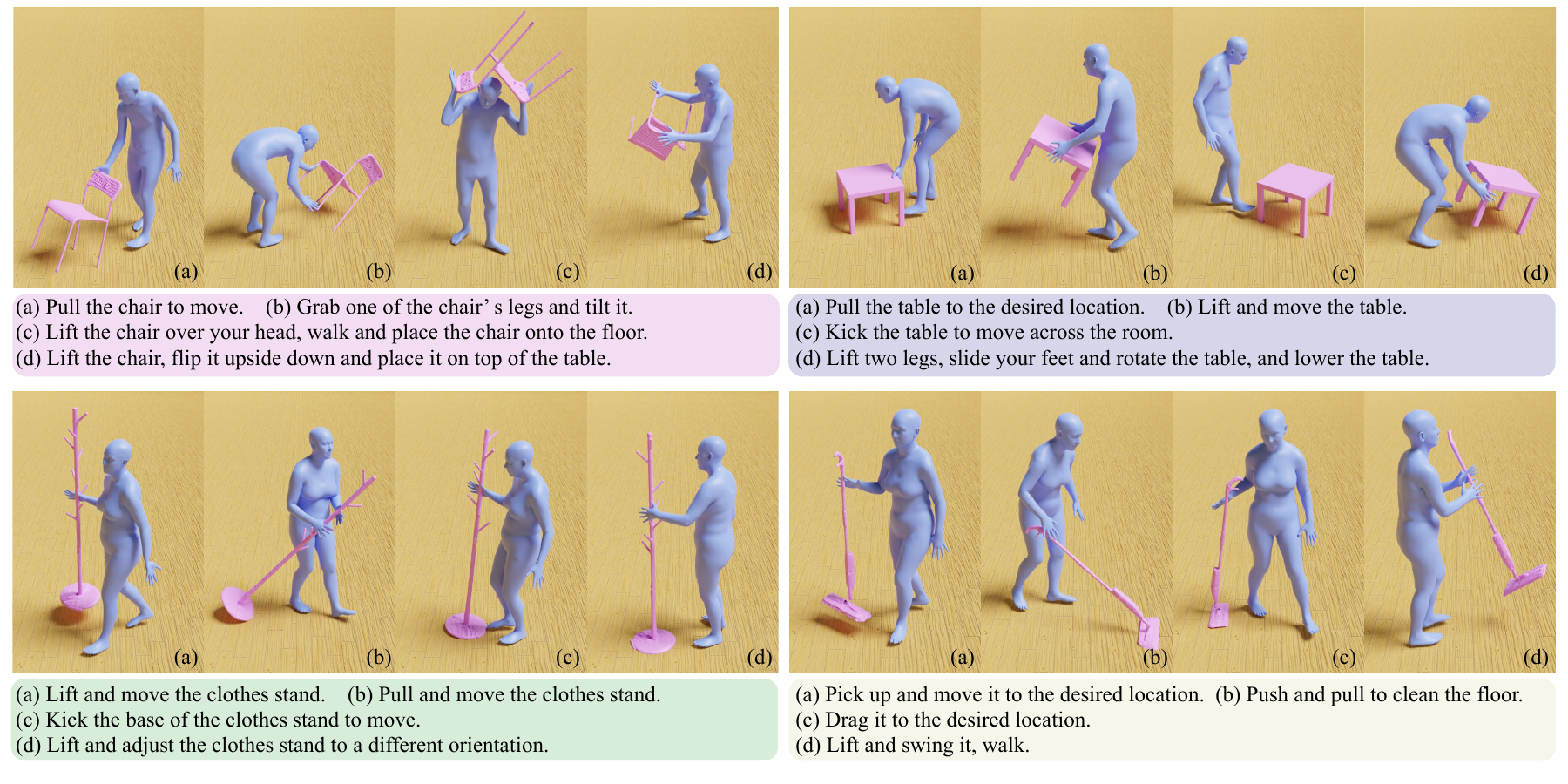}
    \vspace{-8mm}
    \caption{Selected language descriptions used during our mocap sessions.}
    \label{fig:language}
\end{figure*}

\paragraph{Apply Hand Contact Constraints.} 
The hand joint positions generated in the initial stage may not always be precise. They may occasionally deviate from the object, resulting in perceived non-contact at certain time steps. To mitigate this, we propose a post-processing strategy based on the observation that human hands typically maintain consistent relative positions with respect to objects during contact.

Given a sequence of hand joint positions $\bm{H}_1, \bm{H}_2, ..., \bm{H}_T$, we begin by computing the minimum distance from the hand joints to the corresponding object mesh $\bm{V}_1, \bm{V}_2, ..., \bm{V}_T$ at each time step, denoted as $d_1, d_2, ..., d_T$. We then traverse the sequence $d_1, d_2, ..., d_T$ starting from the first frame. We set an empirical contact threshold $\text{th} = 0.03$ and record a specific time step $k$ where $d_k < \text{th}$.

Next, we calculate the difference vector $\bm{p} = \bm{H}_k - \bm{V}_k^{i}$ at step $k$, where $\bm{V}_k^{i}$ is the nearest neighbor vertex of the hand joint on the object mesh. The difference vector $\bm{p}, \bm{p} \in \mathbb{R}^3$, is then used to compute updated hand joint positions in subsequent time steps. We denote the object rotation sequence as $\bm{R}_1, \bm{R}_2, ..., \bm{R}_T$, and for $t > k$, we compute the updated hand joint position as $\hat{\bm{H}}_{t} = \bm{V}_t^{i} + \bm{R}_t\bm{R}_{k}^{-1}\bm{p}$. This ensures the generated hand joint positions maintain a realistic, consistent contact with the object across the entire sequence. From the input object geometry, the first stage determines the joint positions for both the left and right hands. If the positions of both hands are in close proximity to the object, it results in a two-handed manipulation. If not, a single-handed manipulation is established. Specifically, close proximity is determined by computing the Euclidean distance between the hand position and its nearest neighbor points on the object mesh. If this distance is smaller than a predefined threshold (we empirically set the threshold to 0.03), it is inferred that there is contact.


\begin{table}[t]
\centering
\caption{Duration of 15 objects in our dataset.}
\vspace{-5pt}
\begin{tabular}{l*{3}{c}}
\toprule
Object & Large Table & Small Table & Monitor  \\ 
\cmidrule{2-4}
Duration (min) & 37 & 41 & 37    \\
\midrule
Object & Large Box & Small Box & Container  \\
\cmidrule{2-4}
Duration (min) & 40 & 37 & 39 \\
\midrule
Object  &  Wooden Chair & White Chair & Trashcan \\
\cmidrule{2-4}
Duration (min) & 52 & 50 & 34  \\
\midrule
Object & Floor Lamp & Clothes Stand & Tripod  \\
\cmidrule{2-4}
Duration (min) & 35 & 38 & 42  \\
\midrule
Object & Mop & Vacuum & Suitcase  \\
\cmidrule{2-4}
Duration (min) & 42 & 43 & 40  \\
\bottomrule
\end{tabular}
\label{tab:object_duration}
\end{table}

\begin{table*}[ht]
\centering
\caption{Quantitative evaluation on 15 objects.}
\vspace{-10pt}
\begin{tabular}{l*{10}{c}}
\toprule
Method & Hand JPE & MPJPE & MPVPE & $T_{root}$ & $O_{root}$ & Collision \% & FS & $C_{prec}$ & $C_{rec}$ & F1 Score \\
\midrule
GOAL & 49.90 & 15.64 & 21.82 & 34.35 & 0.76 & \textbf{0.12} & \textbf{0.18} & 0.83 & 0.23 & 0.32 \\
OMOMO-single-stage & 26.60 & \textbf{12.07} & \textbf{16.13} & \textbf{17.93} & \textbf{0.47} & 0.19 & 0.38 & 0.78 & 0.42 & 0.51 \\
OMOMO w/o constraints & 24.79 & 12.55 & 16.66 & 18.62 & 0.51 & 0.21 & 0.36 & \textbf{0.83} & 0.58 & 0.64 \\
OMOMO & \textbf{24.01} & 12.42 & 16.67 & 18.44 & 0.50 & 0.22 & 0.38 & 0.82 & \textbf{0.70} & \textbf{0.72} \\
\bottomrule
\end{tabular}
\label{tab:sota_cmp}
\end{table*}

\begin{table*}[ht]
\centering
\caption{Quantitative evaluation on 5 unseen objects.}
\vspace{-10pt}
\begin{tabular}{l*{10}{c}}
\toprule
Method & Hand JPE & MPJPE & MPVPE & $T_{root}$ & $O_{root}$ & Collision \% & FS & $C_{prec}$ & $C_{rec}$ & F1 Score \\
\midrule
GOAL & 53.97 & 15.27 & 20.92 & 40.53 & 0.78 & \textbf{0.06} & \textbf{0.18} & \textbf{0.79} & 0.21 & 0.29 \\
OMOMO-single-stage & 26.06 & \textbf{12.33} & \textbf{16.67} & \textbf{19.49} & \textbf{0.51} & 0.15 & 0.43 & 0.72 & 0.40 & 0.47 \\
OMOMO w/o constraints & 26.15 & 13.25 & 17.77 & 21.55 & 0.53 & 0.16 & 0.45 & 0.76 & 0.44 & 0.52 \\
OMOMO & \textbf{25.12} & 13.06 & 17.60 & 21.19 & 0.53 & 0.17 & 0.43 & 0.74 & \textbf{0.58} & \textbf{0.61} \\
\bottomrule
\end{tabular}
\label{tab:sota_cmp_unseen}
\end{table*}

\paragraph{Generating Full-body Poses from Hand Positions.}
In the second stage, we utilize the same denoising network architecture as in stage one to generate full-body poses from the hand joint positions. The conditions in this stage are the hand joint positions ($\hat{\bm{H}}_1, \hat{\bm{H}}_2, ..., \hat{\bm{H}}_T$) that have been rectified using the contact constraints. The model is trained using human motion data only. 

By integrating these three components, we establish a complete pipeline to generate full-body poses from object motion. This pipeline models the one-to-many mapping from object motion to human poses and ensures that the generated poses maintain realistic contact with the object.

\section{Dataset}
We collected a large-scale high-quality dataset consisting of 3D object geometry, human and object motions. In this section, we elaborate on our object geometry acquisition, motion capture, and data processing. 

\paragraph{Object Geometry Capture}
We selected 15 objects commonly used in everyday tasks, which include a vacuum, mop, floor lamp, clothes stand, tripod, suitcase, plastic container, wooden chair, white chair, large table, small table, large box, small box, trashcan, and monitor. For each object, we filmed a video circling the object and employed Luma~\cite{lumaai} to reconstruct the 3D object geometry from this monocular video. We then utilized Meshlab to manually remove noisy points and downsample object meshes to contain a reasonable number of points for training.

\paragraph{Motion Capture}
We utilized a Vicon system comprised of 12 cameras controlled by Vicon Shogun, which record at a rate of 120 FPS. For each object, we attached 5 markers and captured the object and human motion simultaneously. We invited 17 subjects (13 males, 4 females) to participate in our motion capture sessions. During each mocap session, the volunteer was provided with verbal instructions on how to interact with each object to avoid meaningless interactions. We show some examples of our language guidance in Figure~\ref{fig:language}. Each mocap session lasted approximately 1.5 to 2 hours. The total duration of captured motion for each object is shown in Table~\ref{tab:object_duration}.

\paragraph{Data Processing}
For the object geometry data, we employed a public python library~\cite{mesh2sdf} to compute the SDF for objects. In cases where objects contained noisy SDFs, we used SIREN~\cite{sitzmann2020implicit} to train neural networks and extract the SDF. 

In terms of motion data processing, we used Mosh++~\cite{Loper:SIGASIA:2014, AMASS} to process our raw mocap files and extract SMPL-X model~\cite{smplx} parameters for each sequence. In order to compute object transformations based on marker positions, we initially manually annotate the marker positions on the reconstructed object mesh. Subsequently, we utilize the analytical solution of the orthogonal Procrustes problem to compute the scale, rotation, and translation needed to align the annotated points with the marker positions. Furthermore, we visualize the object and human meshes, and conduct a manual verification on our collected dataset, discarding any sequences that fail to meet our high-quality standard.

\begin{figure}[t]
\begin{center}
    \includegraphics[width=\columnwidth]{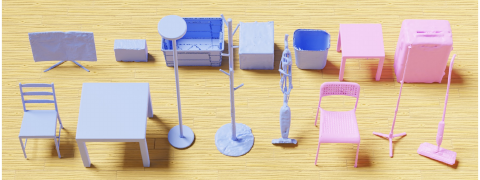}
    \vspace{-6mm}
    \caption{Training objects are annotated in blue, and testing objects are annotated in purple.}
    \label{fig:dataset_objects}
    \vspace{-5mm}
\end{center}
\end{figure}

\begin{figure*}[t!]
    \includegraphics[width=7.0in]{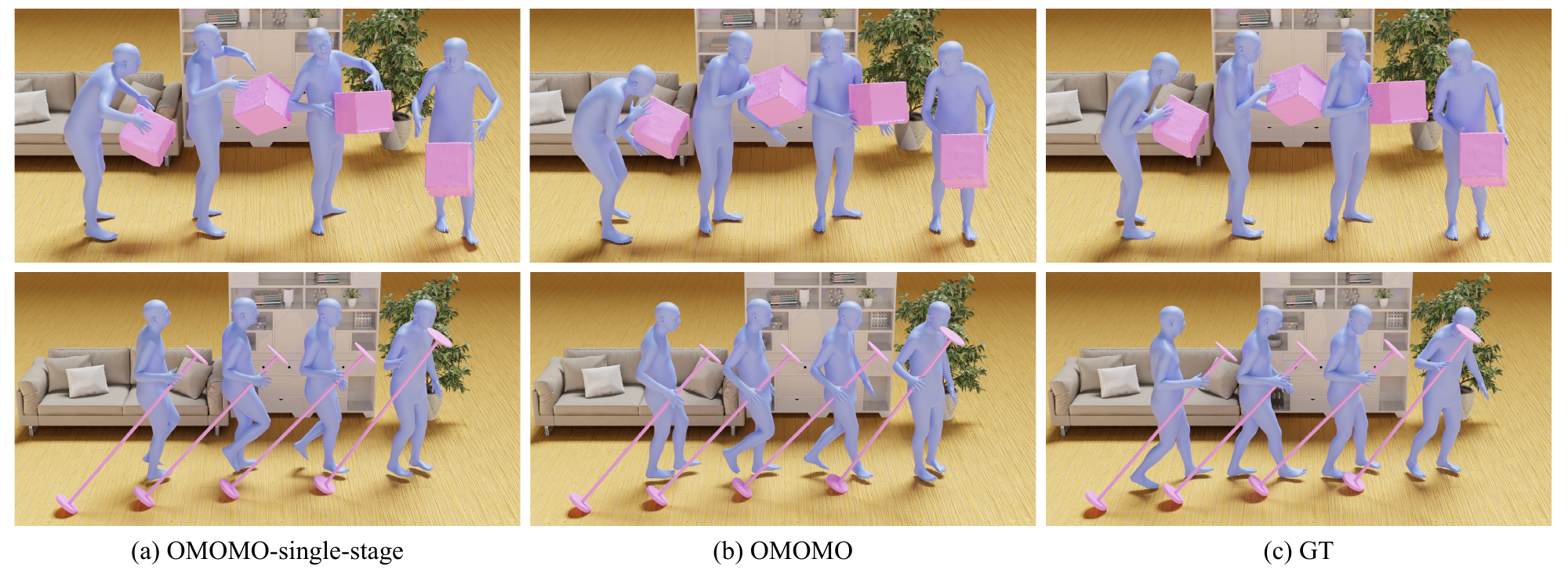}
    \vspace{-4mm}
    \caption{Qualitative Results. We compare our single-stage model, our two-stage model with contact constraints, and ground truth motion. For more qualitative comparisons with GOAL, please watch our supplementary video. }
    \label{fig:results}
    \vspace{-4mm}
\end{figure*}
\section{Experiment}
We first introduce the dataset and evaluation metrics used for this task. Then we describe the chosen baselines and showcase comparisons against them. Additionally, we conduct an ablation study to investigate the effects of hand positions on overall performance. We encourage readers to watch our supplementary video for more qualitative evaluations.

\subsection{Dataset and Evaluation Metrics}
\paragraph{Dataset.} We conduct all experiments using our collected dataset. This dataset consists of motion capture data from 17 subjects, with 15 subjects used for training and 2 subjects for testing. We adopt two data partitioning for evaluation. In the first setting, we use 15 objects for both training and testing. To further evaluate the model's generalization ability to new objects, we divide the 15 objects into 10 for training and 5 for testing as shown in Figure~\ref{fig:dataset_objects}.  

\paragraph{Evaluation Metrics.} We evaluate the synthesized results from two perspectives. Firstly, we compare the generated poses against the ground truth motion data. Additionally, we assess the physical plausibility of the results, considering contact correctness, object penetration, and foot sliding. We detail our evaluation metrics as follows.

\begin{itemize}
    \item \textbf{HandJPE}, \textbf{MPJPE} and \textbf{MPVPE} represent mean hand joint position errors, mean per-joint position errors, and mean per-vertex errors in centimeters $\left(cm\right)$.  
    \item \textbf{$T_{root}$} and \textbf{$O_{root}$} represent the root translation error computed using Euclidean distance in centimeters $\left(cm\right)$ and orientation error defined by the Frobenius norm of the difference between the 3 × 3 rotation matrix $||R_{pred}R^{-1}_{gt} - I||_2$. 
    \item \textbf{FS} represents foot sliding metric and is computed following previous work~\cite{he2022nemf}. 
    \item \textbf{Collision Percentage}. At time step $t$, for $i$th vertex on reconstructed human mesh, we query the object SDF and acquire a signed distance value $d_{t}^{i}$. We use a threshold ($4$cm) to compute collision. If there exists vertices that satisfy $d_{t}^{i} < 0, |d_{t}^{i}| > 4$, we increment the collision count. By traversing the sequence, we can compute the collision percentage. 
  
    \item \textbf{Contact Metrics}. We adopt metrics precision ($C_{prec}$), recall ($C_{rec}$), and F1 score from the object detection task to evaluate contact performance. We first compute the distance between hand positions and object meshes. We empirically set a contact threshold ($5$cm) and use it to extract contact labels for each frame. We perform the same calculation for ground truth hand positions. Then we count true/false positive/negative cases to compute precision, recall, and F1 score.           
\end{itemize}

\subsection{Evaluations}
\paragraph{Baselines.}
Since no existing work specifically addresses the task of object motion-guided human motion synthesis, we adapt a prior work GOAL~\cite{taheri2022goal} on object-reaching motion synthesis as our baseline. GOAL proposed an autoregressive model that predicts future 10 frames conditioned on past 5 poses, hand distance between the current frame and the target goal frame, and the BPS representation which encodes hand-to-object distance at the target frame. In our problem setting, the input is a sequence of object geometry that guide the motion generation instead of a single target frame in GOAL. Thus, we make changes to the input features and use the next frame as the target frame. Specifically, the input features in our modified version consist of the past 5 poses, and the BPS representation that encodes the distance features between the current human mesh and the object mesh at the next frame. 
We use their default model setting which consists of four learning blocks. Each block contains a set of MLPs. The output dimension for each block is 2048, 1024, 1024, 2048 respectively.

\begin{figure}[t]
\begin{center}
    \includegraphics[width=0.95\columnwidth]{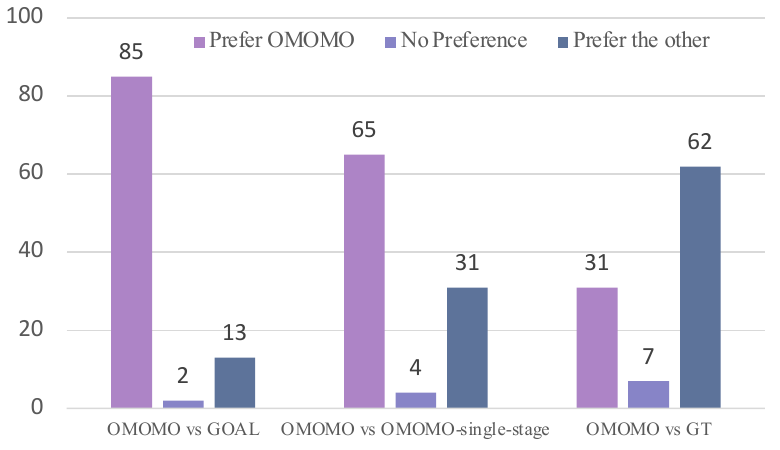}
    \vspace{-3mm}
    \caption{Human Perceptual Study.}
    \label{fig:human_study}
    \vspace{-4mm}
\end{center}
\end{figure}

\paragraph{Implementation Details.}
Our denoising network in OMOMO-single-stage, stage 1 model of OMOMO, and stage 2 model of OMOMO all consist of 4 self-attention blocks with 4 attention heads. The dimension of key, query, and value in the transformer architecture is 256. The output dimension of each layer is 512.  
Our implementation uses PyTorch~\cite{NEURIPS2019_9015}. For training stage 1 and stage 2, we both use Adam optimizer~\cite{kingma2014adam} and start the training with a learning rate 0.0002. The training takes about 18 hours to converge for both stage 1 and stage 2 using a single NVIDIA Titan RTX GPU.

\begin{figure*}[t!]
    \includegraphics[width=7.0in]{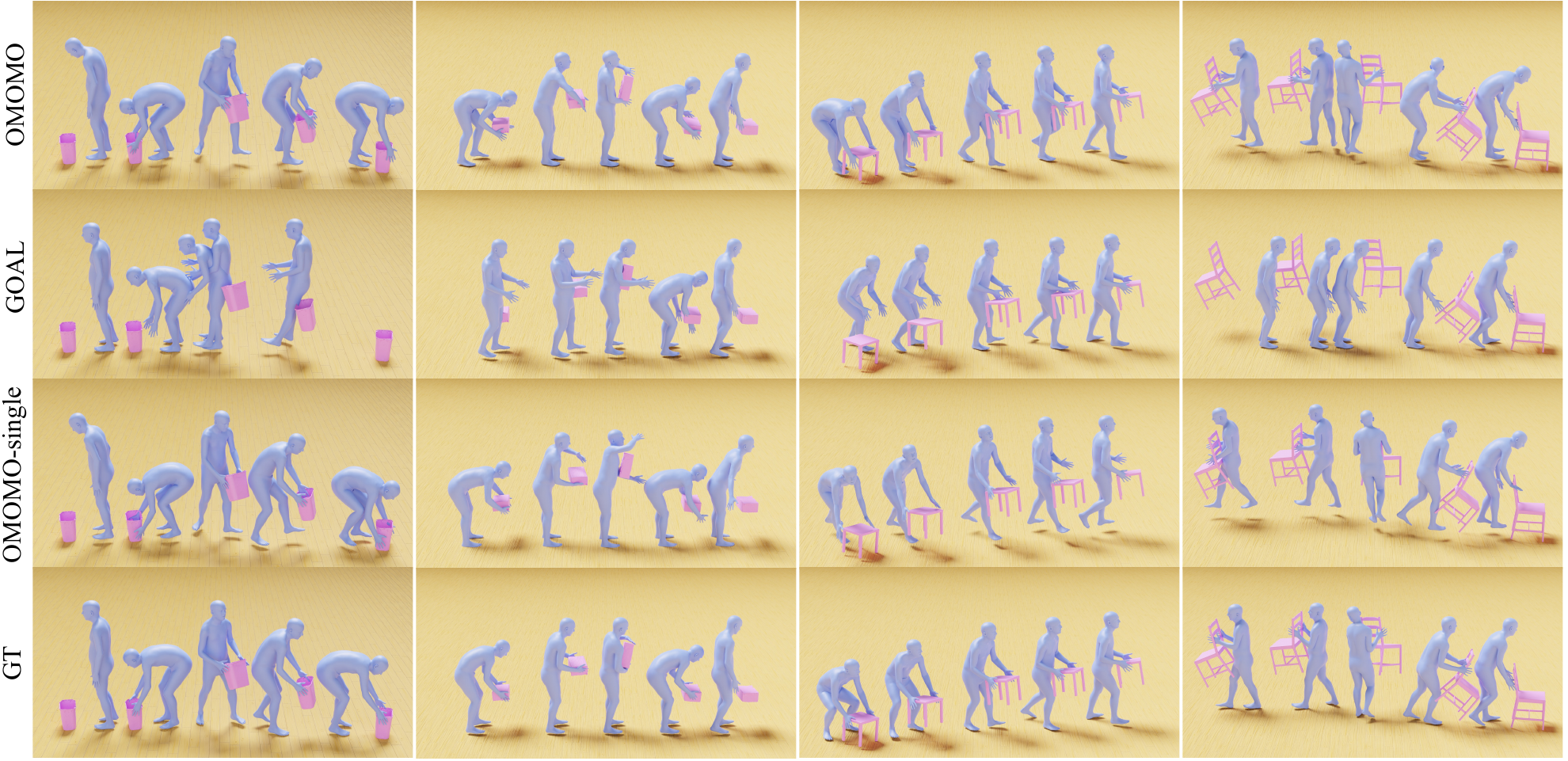}
    \vspace{-2mm}
    \caption{Examples of the generated motion sequences in human perceptual study.}
    \label{fig:more_human_study_results}
    \vspace{-4mm}
\end{figure*}

\begin{figure*}[t!]
    \includegraphics[width=7.0in]{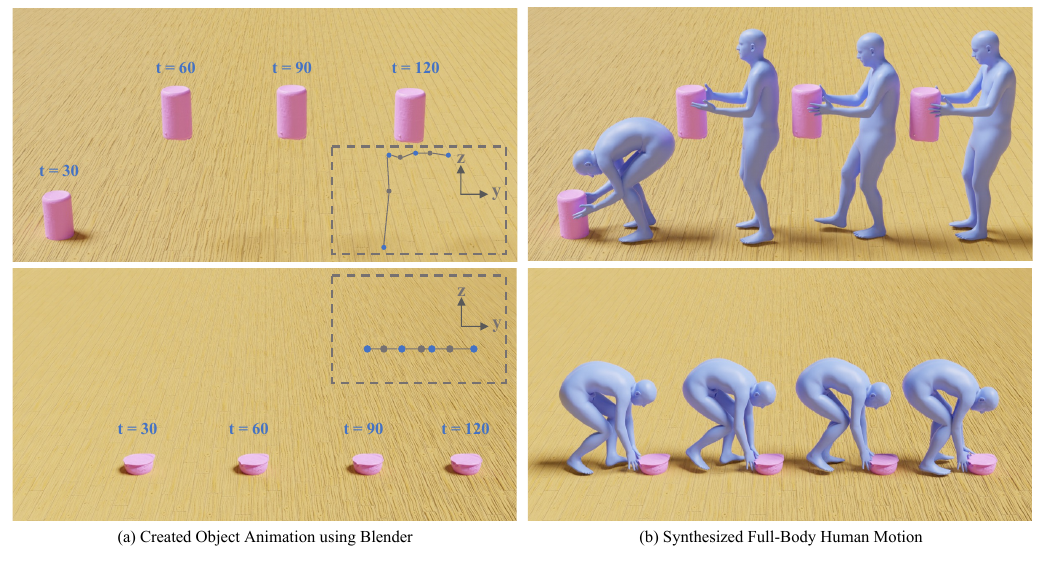}
    \vspace{-5mm}
    \caption{We use Blender to create keyframes for objects with an interval of 15 frames and obtain a sequence of object geometry as input to our pipeline. We show the object geometry every 30 frames and the trajectory of keyframes on yz plane in (a). (b) shows the synthesized human motion.}
    \label{fig:manual_application}
    \vspace{-3mm}
\end{figure*}

\begin{figure*}[ht!]
    \includegraphics[width=7.0in]{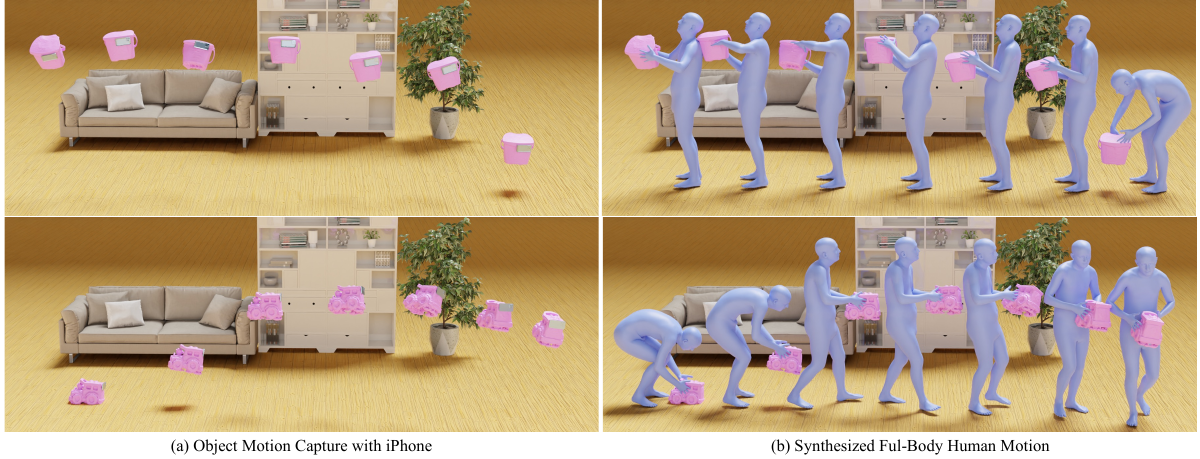}
    \vspace{-3mm}
    \caption{Application. We mount an iPhone on an object (shown in (a)) and use iPhone ARKit to capture object motion. (b) shows the synthesized human motion.}
    \label{fig:application}
    \vspace{-3mm}
\end{figure*}

\paragraph{Results.}
Since our approach is based on conditional diffusion, there can be multiple plausible generation results given the same object motion. To make a quantitative comparison, we sample 20 times for the same object motion input and select the one with the smallest MPJPE. We show quantitative evaluations in Table~\ref{tab:sota_cmp} and Table~\ref{tab:sota_cmp_unseen} for two different data splits. One splits training and testing on all 15 objects. The other one uses 10 objects for training and the other 5 unseen objects for testing. For each configuration, there is only one random seed used to train our model, and the statistics were computed using a single model. Note that our training is not sensitive to random seeds. We outperform baselines in both settings. In particular, OMOMO has superior results in terms of contact evaluations compared to the other two OMOMO variants, which demonstrate the effectiveness of our two-stage design and contact constraints. Note that the reason for the smaller collision percentage in GOAL is that the character in the baseline results often does not attempt to manipulate the object at all, hence the low collision percentage. As for smaller FS scores, we observed that the feet position in the baseline results usually drifts above the floor which will not be counted as foot sliding according to the foot sliding metric. In addition, it is worth mentioning that applying the contact constraints to GOAL is not straightforward. Since GOAL predicts all the joints’ rotations, it requires inverse kinematics to rectify the human pose based on the corrected hand positions.

We also showcase qualitative results in Figure~\ref{fig:results}. OMOMO contains better contacts compared to the setting without hand joint positions as an intermediate representation, as evidenced by Figure \ref{fig:results} and higher contact F1 scores. For more qualitative comparisons, please watch our supplementary video.

\paragraph{Human Perceptual Study}
We further conduct a human perceptual study to complement the evaluations. The goal is to evaluate the motion quality and contact realism. We random sample 100 generated sequences for each approach including OMOMO, OMOMO-single-stage, GOAL, and ground truth, covering all 15 objects. We show some generated results of each approach in Figure~\ref{fig:more_human_study_results}.  We compare OMOMO and the other three settings and totally form 300 pairs for evaluation. For each question, we ask amazon mechanical turk workers which sequence looks more natural and interacts with objects more realistically. Each question is evaluated by 20 different workers ( Figure~\ref{fig:human_study}). 

We show that our OMOMO clearly outperforms the baseline GOAL and OMOMO-single-stage. And when compared with ground truth, 31\% preferred our results (the upper bound would be 50\%). It is worth noting that our results of OMOMO are produced via a single forward pass, without any optimization or post-processing for the full-body poses. Therefore, certain artifacts such as penetration may be produced in the generated motion, which results in ground truth motion is preferred in some sequences. 

\subsection{Ablation Study}
To investigate the effects of hand positions on our overall performance, we compare the full-body human poses generation results that use the predicted hand joint position as input (OMOMO) and ground truth hand positions as input (OMOMO-GT). In Table~\ref{tab:ablation_study}, we show that the synthesis results can be further improved by feeding more accurate hand joint positions.

\begin{table}[t]
\centering
\vspace{5pt}
\caption{Ablation Study. $\ast$ represents the setting that tests on unseen objects.}
\vspace{-5pt}
\begin{tabular}{l*{6}{c}}
\toprule
Method & MPJPE & $T_{root}$ & $C_{prec}$ & $C_{rec}$ & F1 Score \\
\midrule
OMOMO & 12.42 & 18.44 & 0.82 & 0.70 & 0.72  \\
OMOMO-GT & \textbf{7.01} & \textbf{10.08} & \textbf{0.89} & \textbf{0.77} & \textbf{0.79}  \\
OMOMO$^{\ast}$ & 13.06 & 21.19  & 0.74 & 0.58 & 0.61   \\
OMOMO$^{\ast}$-GT & \textbf{7.73} & \textbf{11.08} & \textbf{0.80} & \textbf{0.64} & \textbf{0.67}  \\
\bottomrule
\end{tabular}
\label{tab:ablation_study}
\end{table}

\subsection{Test on Manually Animated Object Trajectory}
We further evaluated our pipeline using manually crafted animations of previously unseen objects. In this process, we began by reconstructing the 3D geometry of the object with the aid of Luma~\cite{lumaai}. Once reconstructed, the 3D object was imported into Blender. Within Blender, we manually established keyframes at 15-frame intervals. Based on these keyframes, Blender then produced a complete object motion sequence. This sequence, exported from Blender, served as the input for our OMOMO. The resulting outputs are shown in Figure~\ref{fig:manual_application}.

\begin{figure}[t]
\begin{center}
    \includegraphics[width=\columnwidth]{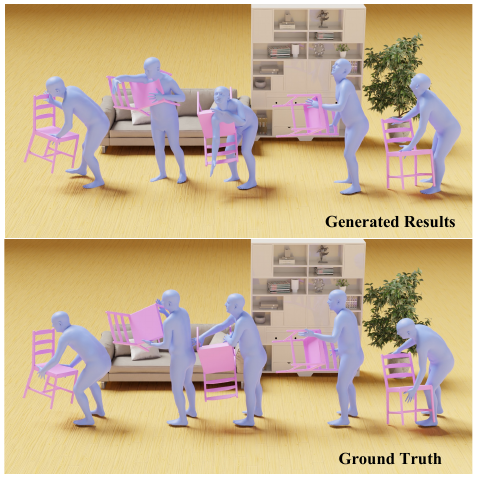}
    \vspace{-6mm}
    \caption{Limitations. Our contact constraint cannot produce generations that involve intermittent contacts with the object. From top to bottom, we show the generated results and corresponding ground truth motion. In the generation results, the hand positions are processed to be fixed on the object, which introduces implausible human motions penetrating with objects.}
    \label{fig:failure_case}
    \vspace{-5mm}
\end{center}
\end{figure}
\section{Application}



We introduce our novel approach to capturing human motion interacting with objects using a single smartphone attached to the object. Specifically, we mount an iPhone XR on the target object and ask the subject to interact with the object while the iPhone camera is filming the environment. We leverage the API ARWorldTrackingConfiguration provided by iPhone ARKit to extract camera poses. This feature is based on visual-inertial odometry techniques that combine visual information and sensor information to estimate accurate camera pose in the world coordinate system. Since the camera is rigidly mounted on objects, we can derive object motion from camera poses. Similar to the data collection process, we film a video and use Luma~\cite{lumaai} to reconstruct 3D geometry of the target object. From a sequence of object-moving geometries, we can generate full-body human poses with our proposed pipeline. We showcase some results in Figure~\ref{fig:application}. Note that these objects are not used during model training.

\section{Conclusion}
In summary, we presented a novel approach for synthesizing human motion guided by moving objects. Specifically, we proposed a framework based on a two-stage paradigm to enforce contact constraints, demonstrating its effectiveness in generating realistic human motions in interaction. Moreover, we introduced a novel application that enables capturing human interaction motion using a smartphone only. To facilitate the research on human-object interactions, we also introduced a large-scale dataset consisting of 3D object geometry, high-quality object motion, and human motion. 

\paragraph{Limitations.} Our current dataset falls short of accurately representing dexterous hand movements, which often result in implausible hand motions. A promising avenue for future research would be incorporating hand priors and optimization techniques, enhancing the realism of hand motions in our full-body pose generations. Furthermore, the contact constraints in our current framework cannot effectively address scenarios with intermittent contacts with the object as shown in Figure~\ref{fig:failure_case}. This could be addressed by identifying and predicting contact states to enable the generation of more complex, long-term manipulation with the objects. Lastly, while our methodology is based on kinematics, future efforts could benefit from integrating physics-based components to mitigate the occurrence of artifacts.

\begin{acks}
This work is in part supported by the Wu Tsai Human Performance Alliance at Stanford University, the Stanford Institute for Human-Centered AI (HAI), NSF CCRI 2120095, ONR MURI N00014-22-1-2740, the Toyota Research Institute (TRI), and Meta. 
\end{acks}
\bibliographystyle{ACM-Reference-Format}
\bibliography{references}

\end{document}